\documentclass[sigconf]{acmart}
\AtBeginDocument{%
  }

\setcopyright{rightsretained}
\copyrightyear{2026}
\acmYear{2026}
\acmDOI{XXXXXXX.XXXXXXX}
\acmConference[ICAIL 2026]{Make sure to enter the correct
  conference title from your rights confirmation email}{June 03--05,
  2026}{Singapore}
\acmISBN{979-8-4007-1939-4/25/06}

\usepackage{enumitem}
\usepackage[most]{tcolorbox}
\usepackage{listings}
\usepackage{xcolor}
\usepackage{placeins}   
\usepackage{caption}
\usepackage{float}

\usepackage{amssymb}
\usepackage{booktabs}
\usepackage[table,xcdraw]{xcolor}
\usepackage{hyperref}

\usepackage{subcaption}

\tcbuselibrary{breakable}

\lstdefinestyle{promptjson}{
  basicstyle=\ttfamily\tiny, 
  columns=fullflexible,
  keepspaces=true,
  showstringspaces=false,
  breaklines=true,
  breakatwhitespace=false,
  tabsize=2,
  upquote=true
}
\newtcblisting{promptbox}{
  enhanced,
  breakable,
  colback=gray!10,
  colframe=black!30,
  boxrule=0.4pt,
  arc=2pt,
  left=1.5mm, right=1.5mm, top=1mm, bottom=1mm,
  listing only,
  listing options={style=promptjson}
}

\begin{document}

\title{LP-Eval: Rubric and Dataset for Measuring the Quality of \\Legal Proposition Generation}

\author{Shanshan Xu}
\email{shanshan.xu@di.ku.dk}
\affiliation{%
  \institution{University of Copenhagen}
  \city{Copenhagen}
  \country{Denmark}
}

\author{Johan Lindholm}
\email{johan.lindholm@umu.se}
\affiliation{%
  \institution{Umeå University}
  \city{Umeå}
  \country{Sweden}
}

\author{Amogh Raina}
\email{amogh.raina@di.ku.dk}
\affiliation{%
  \institution{University of Copenhagen}
  \city{Copenhagen}
  \country{Denmark}
}

\author{Henrik Palmer Olsen}
\email{henrik@jur.ku.dk}
\affiliation{%
  \institution{University of Copenhagen}
  \city{Copenhagen}
  \country{Denmark}
}

\author{Daniel Hershcovich}
\email{dh@di.ku.dk}
\affiliation{%
  \institution{University of Copenhagen}
  \city{Copenhagen}
  \country{Denmark}
}
\renewcommand{\shortauthors}{Trovato et al.}

\renewcommand*{\figureautorefname}{Fig}
\renewcommand{\tableautorefname}{Tab}
\renewcommand{\appendixautorefname}{Appendix}
\renewcommand{\sectionautorefname}{\S}
\renewcommand{\subsectionautorefname}{\S}
\renewcommand{\subsubsectionautorefname}{\S}

\newcommand{\appendixurl}{https://github.com/sxu3/LP-Eval/raw/main/LP-Eval_appendix.pdf}
\newcommand{\appendixref}[1]{\href{\appendixurl}{Appendix~#1}}
\newcommand{\appendixlink}[1]{\href{\appendixurl}{#1}}

\begin{abstract}
Legal proposition generation is central to legal reasoning and doctrinal scholarship, yet remain under-examined in Legal NLP. This paper investigates the automatic generation and evaluation of legal propositions from decisions of the Court of Justice of the European Union using large language models (LLMs). We introduce LP-Eval, a three-step evaluation rubric co-designed with legal experts that decomposes legal proposition quality into formal validity and substantive dimensions. Using this rubric, we release a dataset of two experts' annotations for 100 LLM-generated legal propositions. Our results show that LLMs can generate predominantly well-formed and high-quality propositions, while expert evaluations reveal higher quality for propositions derived from well established cases than from recent ones. We further examine LLMs as evaluators and find that rubric-guided LLM judgments align more closely with expert assessments than direct overall scoring, but remain insensitive to finer-grained distinctions captured by human experts.

\end{abstract}



\received{20 February 2007}
\received[revised]{12 March 2009}
\received[accepted]{5 June 2009}

\maketitle

\section{Introduction}

Legal work, whether in public administration, private corporations, solicitors or barristers offices or courts, centers around determining what the law `is'. Only if we know what the law is, can we apply it to the case we have before us. Lawyers, then, constantly have to state what the law is. Such statements, often called statements on `valid law', articulate the content of general legal rules and principles in some jurisdiction. Such claims, which we in this paper refer to as \textit{Legal Propositions} (LPs), are statements on legal norms (rules or principles, i.e. they are normative and not factual statements), related to their scope, conditions, or consequences. LPs are foundational to doctrinal legal scholarship: the central aim of legal textbooks and legal commentary (often published as journal articles) is to state what the law is by formulating and refining such propositions. 

\begin{figure}

\noindent
\begin{subfigure}{\linewidth}
\begin{tcolorbox}[
    width=0.94\linewidth,
    colback=black!5,
    colframe=black!75,
    title=Automatic Legal Proposition Generation,
    fonttitle=\bfseries]
\footnotesize
\textbf{Paragraph from CJEU Decision:}

\textit{That article therefore lays down strict conditions for the adaptation, by the competent authority of the executing State, ... under Article 8(1) of that framework decision, to recognise a judgment which has been forwarded to it and to execute the sentence ...}

\textbf{Citation Paragraphs:}

\textit{In particular, Article 8 of Framework Decision 2008/909 ...}

\noindent\rule{\linewidth}{0.05pt}

\textbf{Legal Proposition generated by GPT\textsubscript{oss}:}

Article 8(1) of Framework Decision 2008/909 imposes a general duty on the executing State to recognise and execute a forwarded judgment ...
\end{tcolorbox}
\end{subfigure}

\begin{subfigure}{\linewidth}
\begin{tcolorbox}[
    width=0.94\linewidth,
    colback=black!5,
    colframe=black!75,
    title=Expert Annotation based on LN-Eval Rubric,
    fonttitle=\bfseries]

\small\textbf{Formal Components:} Stance \checkmark~Object \checkmark~Specification \checkmark~\textit{(VALID)}

\vspace{0.01cm}
\small\textbf{Quality Dimensions:}
\begin{tabular}{@{}l@{\quad}c@{}}
Source Independence & 2/3 \\
Fact Independence & 3/3 \\
Conciseness & 3/3 \\
Generality & 3/3 \\
Fidelity & 3/3 \\
\end{tabular}

\textbf{Overall:} 3/3~\textit{(Excellent)}

\end{tcolorbox}
\end{subfigure}

\caption{Example of legal proposition generation and expert annotation based on LN-Eval rubric}
\label{fig:lp_generation_example}
\end{figure}

Large-scale collections of legal propositions would have substantial practical value. Legal propositions are lawyers’ primary unit of legal normativity; without them, legal analysis and knowledge organization become markedly more difficult. Legal propositions are today still exclusively manually constructed, either by legal academics when they write textbooks or legal commentary or by legal practitioners when they write pleas or decisions. For example, in the U.S., leading providers of legal information use human editors to generate legal propositions for court decisions, which they refer to as Headnotes \cite{eshelman_history_2018}.

Automating legal proposition generation, the task of producing legal statements based on case decision paragraphs and citation contexts (see upper panel of \autoref{fig:lp_generation_example}), has the potential to improve both the coverage and quality of a wide range of LegalTech applications and downstream legal reasoning pipelines. Despite its practical importance, automatic legal proposition generation has received comparatively limited attention in prior Legal NLP and LegalTech research. Much prior work on legal reasoning has focused either on the sources of law (legislation and prior decisions) e.g., precedent cases retrieval \cite{t-y-s-s-etal-2024-ecthr}, based on the narrative of case fact (e.g., judgment outcome and related classification tasks \cite{chalkidis-etal-2019-neural}. Legal proposition generation is complementary to these lines of work: it targets the interface between “raw” legal sources and the fact patterns of new disputes by producing normative statements that can be invoked, combined, and applied in subsequent analysis. As such, it sits at the \textit{nexus} between legal authority and factual context, offering a potentially useful intermediate representation for both human and machine legal reasoning.

A key challenge in legal proposition generation is evaluation. As an open-ended task, there is rarely a single “gold” reference: multiple outputs may be valid, while small wording differences can alter legal meaning. Consequently, n-gram–based metrics such as BLEU \cite{papineni-etal-2002-bleu} and ROUGE \cite{lin-2004-rouge} are ill-suited, as they capture surface overlap rather than legal content. Another challenge is grounding: propositions must be faithful to authoritative sources. Recent work explores \textit{LLM-as-judge} for evaluation \cite{li2024llmsasjudges}, but this approach remains vulnerable to hallucinations \cite{li-etal-2024-dawn}, such as accepting unsupported claims or fabricated citations \cite{dahl2024large}.

These concerns have motivated recent NLP work in rubric-based evaluation protocols that decompose quality into human-designed explicit, structured criteria (e.g., faithfulness and conciseness) \cite{hashemi2024llmrubric}, rather than a single holistic score. For instance, TN-Eval \cite{shah-etal-2025-tn} develops a rubric and associated protocols to assess the quality of behavioral therapy notes and reports improved reliability and interpretability compared to traditional evaluation approaches. 

Motivated by these challenges, we investigate the ability of LLMs to generate and evaluate legal propositions from decisions of the Court of Justice of the European Union (CJEU). To our knowledge, this is the first study in legal NLP to systematically examine both the automatic generation and evaluation of legal propositions. Our contributions are:\footnote{Our dataset, code and appendix is available at \href{https://github.com/sxu3/LP-Eval}{https://github.com/sxu3/LP-Eval}}
1) We introduce LP-Eval, a three-step rubric co-designed with legal experts to support structured and consistent evaluation of generated legal propositions.
2) We release the LP-Eval dataset, consisting of expert annotations for LLM-generated legal propositions following the LP-Eval guidelines.
3) Through quantitative experiments, we show that LLMs can act as both generators and evaluators of legal propositions, though with notable limitations: experts rate propositions from well established cases higher than those from recent cases, while LLM-based evaluators, despite improved alignment when using LP-Eval, fail to capture such finer-grained distinctions.

\section{Task Definition}
We examine LLMs’ ability to generate and evaluate LPs using decisions of the CJEU. Our experiments comprise two tasks: \emph{LLM as generator} and \emph{LLM as evaluator}.

\noindent\textbf{LLM as generator:}
We prompt off-the-shelf LLMs to generate LPs using an expert-designed template that provides instructions of the task, definition of LP, focus paragraph, and its cited context. \autoref{sec:llm-generattion} offeres details of the prompt and implementation. Two legal experts evaluate the generated propositions using the LP-Eval rubric (\autoref{sec:expert-anno}), which we treat as ground truth for the generation quality.

\noindent\textbf{LLM as evaluator:}
We prompt LLMs to evaluate generated propositions using a template derived from the LP-Eval rubric. Evaluator performance is assessed by measuring agreement between LLM scores and expert annotations. Details of the evaluating prompt and implementation are given in \autoref{sec:llm-evaluation}.

\section{LP-Eval Rubric} 


Building on prior work in rubric design \cite{dawson2017assessment, galvan-sosa-etal-2025-rubriks}, our rubric comprises two elements: (1) required LP \textbf{COMPONENTS}—Stance, Object, and Specification—and (2) quality \textbf{DIMENSIONS} that distinguish stronger from weaker outputs, including Source Independence, Fact Independence, Conciseness, Generality, and Fidelity. The lower panel of \autoref{fig:lp_generation_example} illustrates the rubric. Due to space constraints, a full description is provided in \appendixref{B}.

We developed the rubric through a two-step co-design process with legal experts. First, interviews with two European law professors identified key components and dimensions, which were iteratively refined and summarized into an initial draft. Second, experts applied the rubric to a sample of LLM-generated LPs, followed by feedback to assess its discriminative ability and further refine it. Finally, a computational linguist formalized the evaluation protocol, which we used to develop expert annotation guidelines for human evaluation and a prompt template for LLM-as-judge automatic evaluation.

\section{LP-Eval Dataset}
The dataset consists of 50 paragraphs sampled from 10 CJEU decisions. 
For each paragraph, we include two LLM-generated LPs, manual annotations on the quality of the generated LPs from two legal professionals following the LP-Eval rubric, and corresponding automatic evaluations produced by GPT-oss\cite{agarwal2025gpt}, an open LLM.

\subsection{CJEU Cases Sampling}

We sampled the 10 CJEU decisions from a proprietary database compiled and curated by LawLibrary.AI\footnote{https://www.lawlibrary.ai/} from different official sources. Previous work reveals that LLMs to tend memorize parts of their training data \cite{274574}. A possible concern is therefore that LLMs may generated LPs based on their memorisation of the pretraining data. This is particularly likely when it comes to older and more frequently discussed court decisions. To test for such effects, we selected a combination of \textit{well-established} and \textit{recent} decisions.  The \textit{well-established} group consists of five highly-cited and discussed decisions, spread across time that were manually selected by the legal experts (Full list of the selected cases are available in \appendixref{D}). We selected five paragraphs per case, resulting in a dataset of 50 paragraphs (See \appendixref{D} for sampling details). Descriptive statistics are reported in \appendixref{E}.

\subsection{LLM Legal Proposition Generation}
\label{sec:llm-generattion}
 We employed three models for proposition generation: (1) GPT-OSS: A large-scale open-source model (120 billion parameters) \cite{agarwal2025gpt}. (2) OLMo-3-7B-Instruct: A 7 billion parameter instruction-tuned model \cite{olmo2025olmo}. (3) Saul-7B-Instruct: A 7 billion parameter model specifically trained on legal data \cite{colombo2024saullm}. All models were configured with a temperature of 0.3 to balance between determinism and output diversity. See \appendixref{A} for implementation details and expert-crafted LP generation prompts.
 
\subsection{Expert Evaluation Collection}
\label{sec:expert-anno}
We recruited a final-year law student 
and a research assistant with a PhD in law 
to do the annotation. Both were paid current collective bargaining salaries through their respective universities. They conducted annotations following the LP-Eval annotation protocol. Due to resource constraints, human evaluation was limited to the LPs generated by GPT-OSS and OLMo3, which demonstrated the most promising quality in an initial expert assessment. The annotation process was done on the label studio platform \cite{LabelStudio}.
\\
\noindent\textbf{Inter-Annotator Agreement}
\label{sec:iaa}
We observe a strong positive skew in the experts’ ratings. Only 5 out of 200 generated LPs are labeled as \textbf{INVALID}, due to missing any one of the required formality components (i.e., specification, object, and stance). Across the three quality dimensions rated on a 1–3 Likert scale, the annotation distributions exhibit substantial class imbalance, with the highest rating (“3”) assigned to the majority of items in all dimensions.

Given the pronounced class imbalance, we assess inter-annotator agreement using \textbf{Gwet’s AC1} \cite{gwet2008computing}, consistent with prior work \cite{battisti-etal-2024-advancing}. We prefer AC1 over the more common Cohen’s Kappa due to its robustness against the \textit{prevalence paradox} \cite{feinstein1990high}, which can artificially deflate reliability estimates in skewed distributions despite high observed agreement. We obtained a high agreement score (AC1 = 0.96), indicating strong consistency among annotators and validating the clarity of the LP-Eval Rubric guidelines.


\subsection{Automatic LLM Evaluation}
\label{sec:llm-evaluation}
We also assess proposition quality via an LLM-as-a-judge approach, utilizing GPT-OSS, OLMo3, and Saul as evaluators. All evaluator models were configured with a temperature of 0 to ensure deterministic behavior and maximize reproducibility. The evaluation prompts were constructed using templates derived from the LN-Eval Rubric (see \appendixref{C}).

\subsection{Dataset Statistics}
\label{sec:data-stats}

\begin{table*}[t]
\centering
\resizebox{0.94\linewidth}{!}{%
\begin{tabular}{|l|c|c|c|c|c|}
\hline
 Case Categories& token\_length: paragraph text  & token\_length: GPT generated LP & token\_length: OLMO generated LP & token\_length: Saul-LM generated LP \\
\hline
recent cases & 174.56 ± 50.75 & 52.36 ± 11.86 & 59.12 ± 14.60 & 59.84 ± 25.94 \\
\hline
well established cases & 159.24 ± 68.79 & 45.48 ± 13.02 &  44.84 ± 13.66 & 46.76 ± 16.36 \\
\hline
overall & 166.90 ± 60.32 & 48.92 ± 12.81 & 51.98 ± 15.74 & 53.30 ± 22.46 \\
\hline
\end{tabular}%
}
\caption{Dataset Statistics of the 100 selected CJEU case text paragraphs and generated LPs (mean ± std) }
\label{tab:data_statics}
\end{table*}

\autoref{tab:data_statics} shows that paragraph lengths are comparable across categories, with \emph{recent} cases averaging 174.56 tokens and \emph{well-established} cases 159.24 tokens (std.\ 50.75 and 68.79). Generated LPs are substantially shorter than their source paragraphs for both models: GPT-OSS produces LPs averaging 48.92 tokens, while OLMo-3 produces slightly longer LPs at 51.98 tokens, corresponding to a compression of approximately 29–31\% relative to the average paragraph length (166.90 tokens). The length of generated LPs also varies by model and case category. OLMo-3 generates longer LPs than GPT-OSS for \emph{recent} cases (59.12 vs.\ 52.36 tokens), while their outputs are nearly identical for \emph{well-established} cases (44.84 vs.\ 45.48). Saul-LM produces slightly longer LPs (53.30 tokens) than GPT-OSS (48.92) and is close to OLMo-3 (51.98). The category trend is consistent: Saul-LM generates longer LPs for \emph{recent} cases (59.84) than for \emph{well-established} cases (46.76). 
Notably, Saul-LM shows higher variance in LP length, especially for \emph{recent} cases, which suggests less consistent compression behavior than GPT-OSS or OLMo-3.

\section{Experiments}




\subsection{LLM as LP generator}

\noindent\textbf{Expert Evaluation of LLM-Generated Legal Propositions}\\
We evaluate the quality of LLM-generated LPs through expert assessment following our LP-Eval rubric. An LP is considered valid only if it contains all three required components: \emph{Stance}, \emph{Object}, and \emph{Specification}. Across the dataset, experts judged the vast majority of generated LPs to be formally valid. Out of 100 LPs, 95 were rated as \textsc{VALID}. At the formal \textbf{COMPONENT} level, experts consistently identified the presence of both \emph{Stance} and \emph{Specification} in all generated LPs. The five invalid cases all failed due to a missing \emph{Object} component, defined as a normative (rather than factual) statement.  Furthermore, \emph{Object} component was a significant source of inter-annotator disagreement. 
According to the qualitative study from legal experts (see \appendixref{F}), legal norms consist of different subtypes, which may have confused the expert annotators.

 \begin{table}[h!]
\centering
\resizebox{\linewidth}{!}{
\begin{tabular}{|c|c|c|c|c|c|}
\hline
{\color[HTML]{242523} \textbf{fact\_ind}} & {\color[HTML]{242523} \textbf{fidelity}} & {\color[HTML]{242523} \textbf{concise}} & {\color[HTML]{242523} \textbf{src\_ind}} & {\color[HTML]{242523} \textbf{general}} &
{\color[HTML]{242523} \textbf{overall}} \\ \hline
{\color[HTML]{242523} 2.96 ± 0.19}        & {\color[HTML]{242523} 2.95 ± 0.23}       & {\color[HTML]{242523} 2.48 ± 0.62}      & {\color[HTML]{242523} 2.31 ± 0.63}       & {\color[HTML]{242523} 2.65 ± 0.64}      &
{\color[HTML]{242523} 2.50 ± 0.60}     \\ \hline
\end{tabular}
}

\caption{mean and avg of experts annotation. Likert score 1-3}
\label{tab:quality_dimension_avg}
\end{table}


\autoref{tab:quality_dimension_avg} summarizes expert evaluations at Step~2 (quality \textbf{DIMENSIONS}) and Step~3 (\textbf{OVERALL} quality scores). Experts assigned consistently high scores for \emph{Fact Independence} (mean = 2.96) and \emph{Fidelity} (mean = 2.95), indicating that the generated propositions are largely abstracted from case-specific facts and are well supported by the focus paragraph and its surrounding context. \emph{Conciseness} and \emph{source independence} received slightly lower but still positive average scores (means = 2.48 and 2.31, respectively), reflecting occasional redundancy or explicit references to legal sources. The \textbf{OVERALL} quality score averaged 2.5, rated as high on our 3-point Likert scale. Valid propositions frequently received ratings of 3.0 (“Excellent”) or 2.0 (“Satisfactory”), suggesting that while experts prefer more source-independent drafting, they generally regard the generated propositions as competent and legally sound summaries of the underlying norms.

\begin{table}[h!]
\centering
\resizebox{0.95\linewidth}{!}{
\begin{tabular}{|c|c|c|c|c|}
\hline
          & recent\textsubscript{mean} & established\textsubscript{mean} & t\_statistic & p\_value \\ \hline
fact\_ind & 2.95         & 2.98              & -1.13        & 0.26     \\ \hline
fidelity  & 2.94         & 2.97              & -0.91        & 0.37     \\ \hline
general   & 2.57         & 2.73              & -1.77        & 0.08     \\ \hline
concise   & 2.38         & 2.58              & -2.27        & 2e-02 ** \\ \hline
src\_ind  & 2.11         & 2.52              & -4.74        & 4e-06 ** \\ \hline
overall   & 2.35         & 2.66              & -3.79        & 2e-04 ** \\ \hline
\end{tabular}
}
\caption{T-tests comparing LPs generated from \textit{well established} cases to \textit{recent} cases}
\label{tab:ttest_famous}
\end{table}

\noindent\textbf{Quality Difference in Well-Established vs. Recent Cases}\\
We compared expert evaluation scores for generated LPs derived from \textit{well-established} cases against those from \textit{recent} cases. For each quality dimension, we computed the mean expert score per case-prominence group (established\textsubscript{meann} vs. recent\textsubscript{mean}) and performed paired two-sample t-tests to assess differences between the two groups. As reported in \autoref{tab:ttest_famous}, statistically significant differences emerge for \textit{Conciseness}, \textit{Source Independence}, and \textbf{OVERALL} Quality. LPs generated from \textit{well-established} cases were rated as more concise ($p = 0.02$) and more source-independent ($p < 10^{-5}$). These advantages are reflected in the \textbf{OVERALL} quality scores, which are significantly higher for \textit{well-established} cases (mean = 2.66 vs. 2.35, $p < 0.001$). These results suggest that while core properties, such as factual abstraction and contextual alignment, are stable across case types, higher-level qualities related to abstraction and formulation vary with case prominence. A qualitative analysis by legal experts (\autoref{sec:qualitative_study}) suggests these disparities may stem from the variations in inherent structural differences between the case categories.
 
\subsection{LLM as LP Evaluator}
\noindent\textbf{Alignment Between LLM and Expert Evaluations}


Among the three LLMs tested, OLMo and SaulLM assigned maximum scores to all propositions (zero variance), therefore we focus on the analysis of GPT-OSS's evaluation. 
GPT-OSS achieves Gwet's AC1 agreement of 0.91 and 0.93 with the two experts. While slightly lower than the inter-expert agreement 0.94, these values indicate comparable agreement when GPT-OSS follows the LP-Eval protocol. To test the LN-Eval rubric's efficacy, we prompted GPT-OSS in a \textit{one-go} setting: evaluate only the \textbf{OVERALL} quality score, skipping all intermediate rubric steps for formal COMPONENT and quality Dimensions (see \appendixref{C}). Agreement drops to 0.85 (approximately 7\% relative decrease), demonstrating that structured rubric guidance substantially improves alignment with expert judgments.





\noindent\textbf{LLM-Judge Sensitivity to Case Prominence}
We conducted a t-test comparing GPT-OSS evaluations of LPs derived from \textit{well-established} and \textit{recent} cases but found no statistically significant differences between the two groups. In contrast, expert evaluators identified statistically significant distinctions across these case categories. This suggests that while GPT-OSS broadly aligns with experts at an overall level, it is less sensitive to the nuanced qualitative differences captured by human experts.

\section{Related Work}
\noindent\textbf{Legal NLP}
Recent work in computational law has explored various Legal NLP tasks such as legal case retrieval \cite{ma2021lecard},  judgement
prediction \cite{santosh-etal-2022-deconfounding} , legal case summarization \cite{agarwal-etal-2022-extractive}, vulnerability detection \cite{xu-etal-2023-vechr}, and drafting support for legislative text \cite{chouhan2024lexdrafter}. 
Across these research lines, proposition-level representations remain largely implicit in statute-based formulations \cite{holzenberger2020sara} and are only partially captured in case-law settings \cite{santosh2025lecopcr}. Our work targets this by focusing on proposition-level representations derived from case law and assessing whether language models can generate and preserve the normative propositional content that supports downstream legal reasoning.

\noindent\textbf{LLM-as-a-Judge} is increasingly used as a scalable alternative to expert human assessment for generation tasks, but its reliability depends heavily on the evaluation protocol and prompt design \cite{zheng2023judging}. Recent analyses show that zero-shot LLM evaluators can exhibit systematic biases \cite{stureborg2024biased}. 
To reduce ambiguity and improve reproducibility, recent work emphasizes more explicit rubric construction. CheckEval decomposes subjective criteria into Boolean checklist questions, improving reliability relative to single scalar ratings, improving inter-evaluator agreement \cite{lee2024checkeval}. 

\section{Limitations}

Due to computational constraints, we do not perform a causal analysis of pre-training data and its influence on model outputs. Prior work shows that LLMs can reflect biases in their training data \cite{xu2025better}, motivating future work on detecting and mitigating memorized biases. Owing to space limitations, we emphasize quantitative analysis, with detailed qualitative results provided in \appendixref{F}.

Finally, while expert annotations show high overall agreement, disagreements persist on abstract components (e.g., \emph{object}). This is consistent with prior findings on limited inter-expert agreement in legal NLP \cite{xu-etal-2023-dissonance} and imperfect alignment between models and inherently diverse human judgments \cite{xu2024through}. Given the high-stakes legal setting, future work should better incorporate annotation variability to preserve pluralistic human values and support human-centered AI systems \cite{xu2025noise}.




\section{Conclusion}
We investigate the automatic generation and evaluation of legal propositions (LPs), a core yet underexplored unit of legal reasoning. In collaboration with legal experts, we introduce LP-Eval, a rubric-based evaluation framework for assessing both the validity and quality of generated LPs, along with corresponding evaluation protocols for both expert annotation and LLM-as-judge evaluation. We further release the LP-Eval dataset, a curated corpus of CJEU cases and LLM-generated legal propositions annotated by two legal experts following the LP-Eval rubric. Our experiment results show that modern LLMs can generate generally well-formed and high-quality legal propositions, while exhibiting systematic quality differences across case-prominence types. We also find that rubric-guided LLM evaluation aligns more closely with expert judgments than direct overall scoring, but remains insufficiently sensitive to finer-grained distinctions captured by human evaluators.
These findings highlight both the promise and current limitations of LLMs for legal proposition generation and evaluation. 

\begin{acks}
We thank the anonymous reviewers for valuable comments. This paper is supported by the Independent Research Fund Denmark (DFF) ALIKE grant 426000028B.
\end{acks}

\bibliographystyle{ACM-Reference-Format}
\bibliography{main}


\end{document}